\documentclass[conference]{IEEEtran}
\IEEEoverridecommandlockouts
\usepackage{cite}
\usepackage{amsmath,amssymb,amsfonts}
\usepackage{algorithmic}
\usepackage{graphicx}
\usepackage{textcomp}
\usepackage{xcolor}

\usepackage{booktabs}
\usepackage{epsfig}
\usepackage{algorithm}
\usepackage{ntheorem}
\usepackage{multirow}

\usepackage{subfig}

\newtheorem{theorem}{\bf Theorem}
\newtheorem{definition}{\bf Definition}
\newtheorem{lemma}{\bf Lemma}
\newtheorem*{proof}{\bf Proof}
\newcommand*{\QEDB}{\hfill\ensuremath{\square}}

\def\BibTeX{{\rm B\kern-.05em{\sc i\kern-.025em b}\kern-.08em
    T\kern-.1667em\lower.7ex\hbox{E}\kern-.125emX}}
\begin{document}

\title{Cross-View Graph Consistency Learning for Invariant Graph Representations}

\author{\IEEEauthorblockN{1\textsuperscript{st} Jie Chen}
\IEEEauthorblockA{\textit{College of Computer Science} \\
\textit{Sichuan University}\\
Chengdu, China \\
chenjie2010@scu.edu.cn}
\and
\IEEEauthorblockN{2\textsuperscript{rd} Hua Mao}
\IEEEauthorblockA{\textit{Department of Computer and Information Sciences} \\
\textit{ Northumbria University}\\
Newcastle upon Tyne, U.K. \\
hua.mao@northumbria.ac.uk}
\and
\IEEEauthorblockN{3\textsuperscript{th} Wai Lok Woo}
\IEEEauthorblockA{\textit{Department of Computer and Information Sciences} \\
\textit{ Northumbria University}\\
Newcastle upon Tyne, U.K. \\
wailok.woo@northumbria.ac.uk}
\and
\IEEEauthorblockN{4\textsuperscript{nd} Chuanbin Liu$^{\ast}$ \thanks{*Corresponding author}}
\IEEEauthorblockA{\textit{Center for Scientific Research and Development} \\ \textit{in Higher Education Institutes} \\
\textit{Ministry of Education}\\
Beijing, China \\
liuchuanbin@cutech.edu.cn}
\and
\IEEEauthorblockN{5\textsuperscript{th} Xi Peng}
\IEEEauthorblockA{\textit{ College of Computer Science} \\
\textit{Sichuan University}\\
Chengdu, China \\
pengx.gm@gmail.com}
}

\maketitle

\begin{abstract}
Graph representation learning is fundamental for analyzing graph-structured data. Exploring invariant graph representations remains a challenge for most existing graph representation learning methods. In this paper, we propose a cross-view graph consistency learning (CGCL) method that learns invariant graph representations for link prediction. First, two complementary augmented views are derived from an incomplete graph structure through a coupled graph structure augmentation scheme. This augmentation scheme mitigates the potential information loss that is commonly associated with various data augmentation techniques involving raw graph data, such as edge perturbation, node removal, and attribute masking. Second, we propose a CGCL model that can learn invariant graph representations. A cross-view training scheme is proposed to train the proposed CGCL model. This scheme attempts to maximize the consistency information between one augmented view and the graph structure reconstructed from the other augmented view. Furthermore, we offer a comprehensive theoretical CGCL analysis. This paper empirically and experimentally demonstrates the effectiveness of the proposed CGCL method, achieving competitive results on graph datasets in comparisons with several state-of-the-art algorithms.
\end{abstract}

\begin{IEEEkeywords}
graph representation learning, graph structure augmentation, graph consistency learning, link prediction
\end{IEEEkeywords}

\section{Introduction}
\label{sec:intro}
Graph neural networks (GNNs), inheriting the representation learning advantages of traditional deep neural networks \cite{Wang2023RF, Liu2023TL, Chen2023CSRF, Chen2023LRTL, Peng2022XAI}, have become increasingly popular for analyzing graph-structured data \cite{Hassani2020CMRL}. Graph representation learning (GRL) aims to learn meaningful node embeddings, referred to as graph representations, from both graph structures and node features via GNNs \cite{Kipf2017GCN}. Graph representations have been extensively applied in downstream tasks, e.g., link prediction. Link prediction seeks to predict the missing connections between node pairs from an incomplete graph structure. It shows the significant impact of developing applications with graph-structured data \cite{Liu2023GSL}, including citation network analysis \cite{Li2022MGAE}, social network analysis \cite{Li2022MGAE} and recommendation systems \cite{Wu2021SSGL}.

In recent years, many research efforts have been directed toward investigating GNNs for GRL. For example, several classic GNN models have been proposed to learn meaningful graph representations from graph-structured data, such as graph convolutional network (GCN) \cite{Kipf2017GCN}, variational graph autoencoder (VGAE) \cite{Kipf2016VGAE}, graph attention network \cite{Veli2018GAT}, graph sampling and aggregation (GraphSAGE) \cite{Hamilton2017SAGE} and their variants \cite{Tan2023GAE, Li2022MGAE}. These methods have achieved impressive link prediction results. However, the potential of utilizing graph structures for the available graph-structured data has not been fully exploited.

Self-supervised learning has recently emerged as a promising representation learning paradigm for GNNs \cite{Hou2022SMGA, Tsai2020NM}. It can learn latent graph representations from unlabeled graph-structured data by supervision, which is provided by the data itself with different auxiliary learning tasks. Most self-supervised learning-based algorithms fall into two categories: contrastive learning \cite{Chen2023DMC, Li2022TCL, Hassani2020CMRL, You2020GCL} and generative learning \cite{Tan2023GAE, Li2022MGAE, Hou2022SMGA, Cui2020ADE}. For example, Hassani \textit{et al.} \cite{Hassani2020CMRL} presented a contrastive multiview representation learning (CMRL) method that learns graph representations by contrasting encodings derived from first-order neighbors and a graph diffusion module. The feature pairs may have different data distributions under the two different types of augmented views. This may have a significant negative impact on measuring the similarity between positive pairs and the dissimilarity between negative pairs when conducting contrastive learning. Li \textit{et al.} \cite{Li2022MGAE} presented a masked graph autoencoder (MGAE) method that reconstructs a complete graph structure by masking a portion of the observed edges. By randomly masking a portion of the edges, the MGAE method somewhat reduces the redundancy of the graph autoencoder (GAE) in self-supervised graph learning tasks. The randomness of edge masking causes a sampling information loss problem. Consequently, this is an increasing concern regarding the capture of the complementary information between the augmented views of a graph.

\begin{figure*}[!htbp]
\centering
\includegraphics[width=0.6\linewidth]{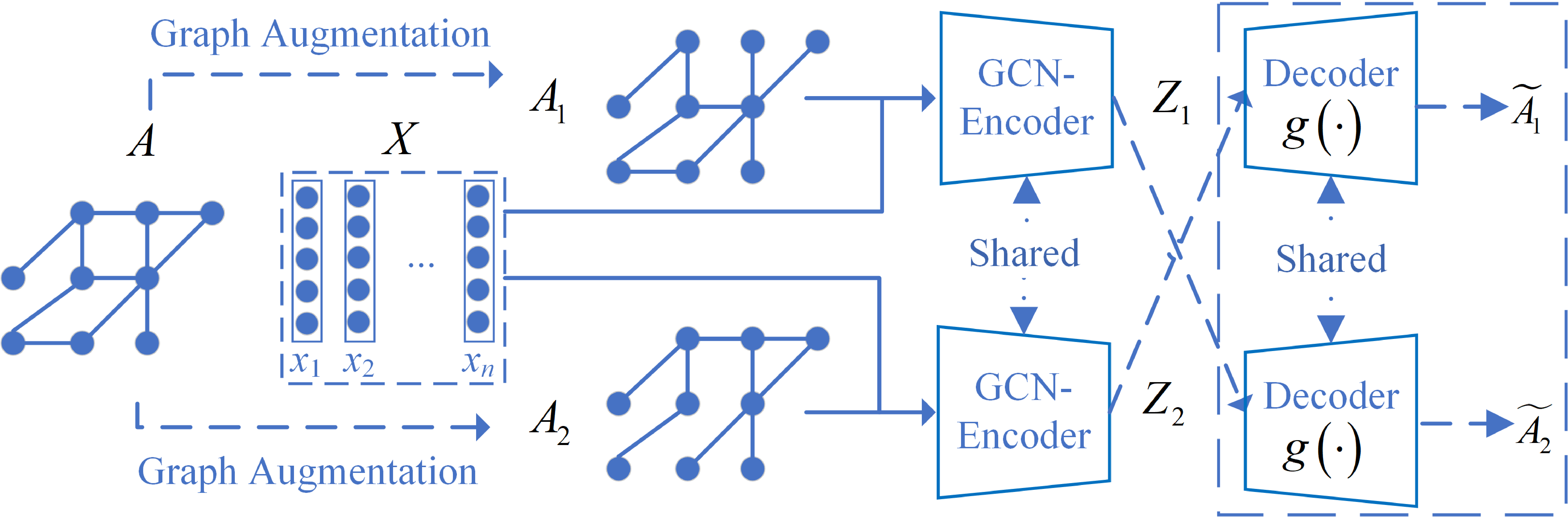}
\caption{Framework of the CGCL model. Each augmented view of a graph structure, ${\mathbf{A}_i}$ $\left( {i = 1,2} \right)$, corresponds to three modules, including an augmented graph structure module, a shared GCN encoder module and a shared cross-view consistency decoder module. $\mathbf{A}$ and $\widetilde {\mathbf{A}}$ represent the incomplete graph structure and predictive graph structure, respectively.}
\label{fig:architecture} 
\end{figure*}

A vast majority of the available contrastive learning-based GRL methods consider their GNN-based models to be independent of different downstream tasks \cite{Tan2023GAE, Zhu2020GCRL}. Recent advances have shown that the optimal data augmentation views critically depend on the downstream tasks involving the visual data \cite{Wang2022MSR, You2020GCL}. The augmented views of the visual data share as little information as necessary to maximize the task-relevant mutual information. Motivated by this, we further investigate how to take full advantage of graph structure information and simultaneously retain the task-relevant information needed by a specific downstream task, i.e., link prediction. This is beneficial for learning invariant graph representations from the augmented views of graph-structured data, which is central to enhancing the ability of a model to produce general graph representations.

In this paper, we propose a cross-view graph consistency learning (CGCL) method that learns invariant graph representations for link prediction. Numerous missing connections are involved in the graph structure. We first construct two complementary augmented views of the graph structure of interest from the remainder of the link connections. Then, we propose a cross-view training scheme to train the proposed CGCL model, which can produce invariant graph representations for graph structure reconstruction purposes. The proposed cross-view training scheme attempts to maximize the consistency information between one augmented view and the graph structure reconstructed from the other augmented view. In contrast with random edge perturbation, node dropping or attribute masking, this approach preserves the valuable information contained in raw graph-structured data during GRL. Moreover, the complementary information in the graph structure can be further exploited by virtue of the cross-view training scheme. In addition, a comprehensive theoretical analysis is provided to reveal the supervisory information connections between the two complementary augmented views.

The key contributions are summarized as follows.
\begin{itemize}
\item{We propose a CGCL model that learns two pairs of cross-view adjacency matrices, which can be considered view-invariant, for link prediction.}
\item{A coupled graph structure augmentation scheme is introduced to construct two complementary augmented views of a graph structure, which help produce more distinct low-dimensional node embeddings.}
\item{A cross-view training scheme is designed to train the CGCL model by maximizing the consistency information between one augmented view and the graph structure reconstructed from the other augmented view.}
\item{Extensive experiments are conducted on graph datasets, achieving competitive results.}
\end{itemize}

\section{Preliminaries}
\label{sec:prelims}
An undirected graph $\mathcal{G}$ is defined as $\mathcal{G} = \left( {\mathcal{V},\mathcal{E}} \right)$, where $\mathcal{V} = \left\{ {{v_1},...,{v_n}} \right\}$ and $\mathcal{E} = \left\{ {{e_1},...,{e_m}} \right\}$ stand for a set of $n$ nodes and a set of $m$ edges, respectively \cite{Kipf2017GCN}. Each node $v_i$ in $\mathcal{V}$ is associated with a corresponding feature $x_i$ $\left( {1 \le i \le n} \right) \in {\mathbb{R}^d}$. An adjacency matrix $\mathbf{A} \in {\mathbb{R}^{n \times n}}$ represents the relationships among the nodes in a graph, where ${A_{ij}} = 1$ indicates that an edge exists between nodes $v_i$ and $v_j$ and vice versa. The degree matrix $\mathbf{D}$ is defined as $\mathbf{D} = diag\left[ {{d_1},{d_2},...,{d_n}} \right] \in {\mathbb{R}^{n \times n}}$, and its diagonal elements are ${d_i} = \sum\nolimits_{{v_j} \in V} {{A_{ij}}}$.

A GCN learns node embeddings for graph-structured data \cite{Kipf2017GCN}. Given an undirected graph $\mathcal{G}$, $\widetilde {\mathbf{A}} = \mathbf{A} + \mathbf{I}$ is the adjacency matrix of $\mathcal{G}$ with an added self-loop, and ${\widetilde {\mathbf{D}}}$ is a diagonal degree matrix ${\widetilde D_{ii}} = \sum\nolimits_j {{{\widetilde A}_{ij}}}$. The formula for the $l$th GCN layer is defined as
\begin{equation}\label{eq:gcn}
\begin{split}
{\mathbf{H}^{(l)}} & = \sigma \left( {{{\widetilde {\mathbf{D}}}^{ - {1 \mathord{\left/
 {\vphantom {1 2}} \right.
 \kern-\nulldelimiterspace} 2}}}{\widetilde {\mathbf{A}}}{{\widetilde {\mathbf{D}}}^{ - {1 \mathord{\left/
 {\vphantom {1 2}} \right.
 \kern-\nulldelimiterspace} 2}}}{\mathbf{H}^{(l - 1)}}{\mathbf{W}^{(l)}}} \right) \\
\end{split}
\end{equation}
where $l$ denotes the $l$th layer, ${\mathbf{W}^{(l)}}$ is a layer-specific learnable weight matrix, ${\mathbf{H}^{(l)}}$ is the node embedding matrix with $\mathbf{H}^{(0)}=\mathbf{X}$, and $\sigma$ is a nonlinear activation function, e.g., ${\rm ReLU}\left(  \cdot  \right) = \max \left( {0, \cdot } \right)$. For a semi-supervised classification task, the weight parameters in the GCN model can be learned by minimizing the cross-entropy error between the ground truth and predictive the labels \cite{Kipf2017GCN}.

\section{Cross-View Graph Consistency Learning}
\label{sec:cgcl}
In this section, we present the proposed CGCL method in detail, which produces invariant graph representations for self-supervised graph learning. These graph representations can be employed on a specific downstream task, i.e., the reconstruction of an incomplete graph structure.

\subsection{Problem Formulation}
Let $\mathbf{X} \in {\mathbb{R}^{n \times d}}$ be a matrix consisting of node features, each row of which corresponds to a node feature. Given an undirected graph $\mathcal{G}$ with an incomplete graph structure and the above node features $\mathbf{X}$, the goal of our work is to learn graph representations, which can be employed to reconstruct the incomplete graph structure for link prediction.

\subsection{Network Architecture}
The proposed CGCL method aims to produce the complete graph structure by predicting the missing connections. Fig. \ref{fig:architecture} provides an overview of the proposed CGCL network architecture, which is composed of three main modules, i.e., an augmented graph structure module, a shared GCN encoder module and a shared cross-view consistency decoder module. The augmented graph structure module is utilized to generate two complementary augmented views of the original graph structure. Inspired by the VGAE \cite{Kipf2016VGAE}, the shared GCN encoder module learns individual graph structures for each augmented view under unsupervised representation learning. The cross-view consistency decoder module produces a predictive graph structure by maximizing the consistency information between one augmented view and the graph structure reconstructed from the other augmented view. With these three modules, CGCL simultaneously learns graph representations from graph-structured data and reconstructs an incomplete graph structure for inferring the missing connections.

\subsection{CGCL Model}

\subsubsection{Motivation}
Considering the two complementary graph structures ${\mathbf{A}_1}$ and ${\mathbf{A}_2}$ shown in Fig. \ref{fig:architecture}, CGCL aims to ensure the consistency of the pairwise matchings between the pairs of nodes in ${\mathbf{A}_1}$ and $\widetilde {{\mathbf{A}_1}}$, where $\widetilde {{\mathbf{A}_1}}$ is produced from the other original structure ${\mathbf{A}_2}$ and the node features.

\begin{definition}[Cross-View Graph Consistency] \label{def:consistency}
Given two augmented graph structures ${\mathbf{A}_1}$ and ${\mathbf{A}_2}$, a generated graph structure $\widetilde {{\mathbf{A}_1}}$ is assumed to be reconstructed by using the node features $\mathbf{X}$ and graph structure ${\mathbf{A}_2}$. Let $A_{ij}^1$ and $\widetilde {A_{ij}^1}$ $\left( {i,j \in \left\{ {1,...,n} \right\}} \right)$ form a pairwise matching chosen from ${\mathbf{A}_1}$ and $\widetilde {{\mathbf{A}_1}}$, respectively. The relationship between the two graph structures ${\mathbf{A}_1}$ and $\widetilde {{\mathbf{A}_1}}$ is said to exhibit cross-view consistency if the following equality holds: $A_{ij}^1 = \widetilde {A_{ij}^1}$ for all $i,j \in \left\{ {1,...,n} \right\}$.
\end{definition}

To reconstruct the incomplete graph structure, CGCL maximizes the consistency information between two random variables $v_1$ and $v_2$ with a joint distribution $p\left( {{v_1},{v_2}} \right)$ corresponding to the pairwise matching variables of a pair of nodes ${a_1}$ and ${a_2}$ in a graph structure ${\mathbf{A}_1}$ and its reconstruction $\widetilde {{\mathbf{A}_1}}$, i.e.,
\begin{equation}\label{eq:cons1}
\begin{split}
\mathop {\max }\limits_f C\left( {{a_1},{a_2}} \right)
\end{split}
\end{equation}
where $C\left( {{a_1},{a_2}} \right) = {\mathbb{E}_{p_{({a_1},{a_2})}}}{a_1}\log {a_2}$, $ {a_i} = f\left( {{v_i}} \right)$ denote random variables, ${v_i} \ge 0, i \in \left\{ {1,2} \right\}$, and $f$ represents a mapping function. According to the data processing inequality for the Markov chains $v_1 \to v_2 \to a_2$ and $a_2 \to v_1 \to a_1$ \cite{Wang2022MSR, Cover2006IT}, we have
\begin{equation}\label{eq:cons2}
\begin{split}
C\left( {{v_1},{v_2}} \right) \ge C\left( {{v_1},{a_2}} \right) \ge C\left( {{a_1},{a_2}} \right).
\end{split}
\end{equation}
Thus, $C\left( {{v_1},{v_2}} \right)$ is the upper bound of $C\left( {{a_1},{a_2}} \right)$. The variable $v_1$ can provide supervisory information for $v_2$ in an unsupervised manner, and vice versa. All supervisory information for one augmented view comes from the consistency information provided by the other augmented view in the context of CGCL. Assuming that the mapping function $f$ has a sufficient graph representation learning ability, we have $C\left( {{v_1},{v_2}} \right) = C\left( {{a_1},{a_2}} \right)$. This indicates that $C\left( {{v_1},{v_2}} \right)$ and $C\left( {{a_1},{a_2}} \right)$ are approximately minimal when the consistency information shared between the two augmented views is sufficient for each other during GRL.

\subsubsection{Coupled Graph Structure Augmentation}
The hidden representation of each node during GRL is determined by both the node itself and its neighbors. The neighbor selection process is influenced by both the graph structure and the node features. To extract sufficient supervisory information from the two augmented views, the straightforward approach is to select distinct neighbor candidate sets for each node. In contrast to the recently proposed graph contrastive learning methods that employ techniques such as node dropping and attribute masking for graph data augmentation, we refrain from performing any augmentation operations on the node features of the graph.

Guided by the task of reconstructing the incomplete graph structure, a portion of the connections is missing in an undirected graph $\mathcal{G}$. Edge dropping is an intuitive graph-structured data augmentation. The choice of different edge dropping ratios often leads to varying results in link prediction tasks. Some combinations of dropping ratios can even perform worse than no augmentation at all. Consequently, determining an optimal dropping ratio for edges becomes a dilemma. We introduce a coupled graph structure augmentation scheme for the original incomplete graph structure. Specifically, we randomly divide the set of edges into two complementary subsets following a particular distribution, e.g., the Bernoulli distribution. The edges of the two subsets are complementary when the directions of the edges are not considered. Furthermore, two undirected augmented views $\mathbf{A}_1$ and $\mathbf{A}_2$ can be constructed after applying the bidirectional order for each pair of nodes in each subset, which are $\mathcal{E}_1$ and $\mathcal{E}_2$, respectively. In particular, these two views may share some edge pairs. This strategy offers two benefits for graph-structured augmentation. First, it directly simplifies the process of selecting the optimal edge-dropping ratio. Second, graph representation learning models with multiple GNN layers tend to make node representations more similar due to their message-passing mechanisms. By providing complementary candidate neighbor sets for each node, these two augmented views help produce more distinct low-dimensional node embeddings.

Given the adjacent matrices of the two complementary augmented views for a graph structure, $\mathbf{A}_1$ and $\mathbf{A}_2$,  two corresponding adjacency matrices can be reconstructed using the GNN-based models \cite{Veli2018GAT, Hamilton2017SAGE, Kipf2017GCN}. Different from previous work \cite{Tan2023GAE, Fang2022GRL, Li2022MGAE}, we focus on reconstructing the cross-view adjacency matrices from $\mathbf{A}_1$ and $\mathbf{A}_2$ based on our specific motivation. Specifically, the cross-view adjacency matrices $\mathbf{\widetilde {A}}_1$ and $\mathbf{\widetilde {A}}_2$ can be reconstructed as follows:
\begin{equation}\label{eq:ff}
\begin{split}
& {\widetilde {\mathbf{A}}}_1 = {f}\left( \mathbf{X},\mathbf{A}_2 \right),\\
& {\widetilde {\mathbf{A}}}_2 = {f}\left( \mathbf{X},\mathbf{A}_1 \right)\\
\end{split}
\end{equation}
where ${f}$ is a mapping function consisting of the shared GCN encoder module and shared cross-view consistency decoder module shown in Fig. \ref{fig:architecture}. ${\widetilde {\mathbf{A}}}_1$ and ${\widetilde {\mathbf{A}}}_2$ represent the prediction results obtained for the incomplete graph structures derived from $\mathbf{A}_2$ and $\mathbf{A}_1$, respectively. Two pairs of adjacency matrices produced from graph-structured data, $\left(\mathbf{A}_1, {\widetilde {\mathbf{A}}}_1\right)$ and $\left(\mathbf{A}_2, {\widetilde {\mathbf{A}}}_2\right)$, are cross-view adjacency matrices that can be considered view-invariant matrices. These pairs of adjacency matrices can be utilized for link prediction.


\subsubsection{Reconstruction of an Incomplete Graph Structure}
The shared GCN encoder module utilizes a GCN as the backbone. For each GCN-based encoder component, the encoder part produces hidden representations of the nodes derived from a graph structure and node features. Without a loss of generality, the GCN-based encoder component is defined as
\begin{equation}\label{eq:fz}
\begin{split}
\mathbf{Z}_v & = {g }\left( {\mathbf{X},\mathbf{A}_{v}} \right) = {\mathop{\rm ELU}\nolimits}\left( {{\mathbf{A}_{v}}\mathbf{X}\mathbf{W}^{(1)}} \right),\\
\end{split}
\end{equation}
where $\mathbf{W}^{(1)} \in {\mathbb{R}^{d \times d_v}}$ is the weight parameters, $d_v$ represents the number of neural units in the hidden layer, ${\rm ELU}\left(  \cdot  \right)$ denotes a nonlinear activation function, and $\mathbf{Z}_v$ represents low-dimensional node embeddings. Thus, the GCN-based encoder component extracts graph-level representations $\mathbf{Z}_1$ and $\mathbf{Z}_2$ for the adjacent matrices of the two augmented views $\mathbf{A}_1$ and $\mathbf{A}_2$, respectively.

Given the low-dimensional node embeddings $\mathbf{Z}_{v}$, we employ the shared cross-view consistency decoder module, which consists of an inner product decoder component and a multilayer perceptron (MLP) component, to obtain the reconstructed adjacency matrices. The inner product decoder component is constructed by
\begin{equation}\label{eq:fip}
\begin{split}
& {\mathbf{H}_{v}} = \sum\limits_{i = 1}^N {\sum\limits_{j = 1}^N {p\left( {{\mathbf{A}_{ij}^{(v)}}|{\mathbf{Z}_i^{(v)}},{\mathbf{Z}_j^{(v)}}} \right)} }
\end{split}
\end{equation}
where $p\left( {{\mathbf{A}_{ij}^{(v)}}|{\mathbf{Z}_i}^{(v)},{\mathbf{Z}_j}^{(v)}} \right) = { \left( {\mathbf{Z}_i^{(v)}} \right)^T{\mathbf{Z}_j^{(v)}}}$ and $v \in \left\{ {1,2} \right\}$. The MLP component is composed of two fully connected layers with computations defined as follows:
\begin{equation}\label{eq:fmlp}
\begin{split}
& \mathbf{\widetilde {A}}_1 = \mathbf{MLP}\left( {{\mathbf{H}_{2}},{\mathbf{W}^{(1,2)}}} \right), \\
& \mathbf{\widetilde {A}}_2 = \mathbf{MLP}\left( {{\mathbf{H}_{1}},{\mathbf{W}^{(1,2)}}} \right) \\
\end{split}
\end{equation}
where a nonlinear activation function, ${\rm ReLU}\left(  \cdot  \right)$, is applied for each layer. $\mathbf{H}_1$ and $\mathbf{H}_2$ represent invariant graph representations that are utilized to predict $ \mathbf{\widetilde {A}}_2$ and $ \mathbf{\widetilde {A}}_1$, respectively. The reconstructed graph structures $\widetilde {\mathbf{A}_1}$ and $\widetilde {\mathbf{A}_2}$ can be obtained using Eq. \eqref{eq:fmlp}.

\subsection{Training Scheme}
\label{sec:training}
We propose a cross-view scheme for training the CGCL model. For the connections between the same pair of nodes, we emphasis their consistency across the augmented view and its corresponding counterpart. Specifically, the difference between the adjacent matrices of the augmented view ${\mathbf{A}_v}$ and the reconstructed graph structure ${\widetilde {\mathbf{A}_v}}$ should be reduced during the training stage $ \left( v \in \left\{ {1,2} \right\} \right)$. We establish two sets of edges to measure this difference. First, we create a set of edges by selecting all edges from $\mathbf{A}_v$ $ \left( v \in \left\{ {1,2} \right\} \right)$, considering each edge as positive. Second, we generate a set of negative edges by randomly sampling unconnected edges from the alternate augmented view. The number of negative edges corresponds to the number of positive edges within the alternative augmented view. By adopting the binary cross-entropy loss, the loss function for the graph consistency $L_v$ between ${\mathbf{A}_v}$ and ${\widetilde {\mathbf{A}_v}}$ is formulated as:
\begin{equation}\label{eq:lossr}
\begin{split}
{L_v} =& \ {loss\left( {\mathbf{A}_v}, {\widetilde {\mathbf{A}_v}}  \right)} \\
 = & -\frac{1}{{{n^2}}}\sum\limits_{i = 1}^n {\sum\limits_{j = 1}^n \left[ \delta \left( {\mathbf{A}_{ij}^v}  \right)  \right. } \log \delta \left( {\widetilde {\mathbf{A}_{ij}^v}} \right)  \\
& + \left( {1 - \delta \left( { \mathbf{A}_{ij}^v} \right) } \right) \left. \log \left( {1 - \delta \left( {\widetilde {\mathbf{A}_{ij}^v} } \right)} \right) \right] \\
\end{split}
\end{equation}
where ${\mathbf{A}_{ij}^v} $ and ${\widetilde {\mathbf{A}_{ij}^v}}$ represent the elements of the $i$th rows and the $j$tth columns of ${\mathbf{A}_v}$ and ${\widetilde {\mathbf{A}_v}}$, respectively. The entire learning procedure of the proposed CGCL method is summarized in Algorithm \ref{alg:dgcl}. Thus, we can obtain the graph representations and the predictive graph structure, simultaneously.

\subsection{Theoretical Analysis}
In this section, we provide a theoretical analysis of our model from the perspective of sufficient supervision information in GRL. The reconstructions of the incomplete graph structures ${\widetilde {\mathbf{A}_1}}$ and ${\widetilde {\mathbf{A}_2}}$ are obtained from the two augmented views. Considering Eq. \eqref{eq:lossr}, the general form of the optimization objective is
\begin{equation}\label{eq:general}
\begin{split}
& loss \left( {\mathbf{A}_v}, \widetilde {\mathbf{A}_v} \right) \\
\end{split}
\end{equation}
where $v \in \left\{ {1,2} \right\}$. According to Eq. \eqref{eq:cons2}, a loss of task-relevant information occurs if the following condition holds, i.e.,
\begin{equation}\label{eq:infoloss}
\begin{split}
C\left( {\mathbf{A}_1},{\mathbf{A}_2} \right) \ge C\left( {\widetilde {{\mathbf{A}_1}},\widetilde {{\mathbf{A}_2}}} \right).
\end{split}
\end{equation}

\begin{algorithm}
\renewcommand{\algorithmicrequire}{\textbf{Input:}}
\renewcommand\algorithmicensure {\textbf{Output:} }
\caption{Optimization Procedure for CGCL}
\label{alg:dgcl}
\begin{algorithmic}
\REQUIRE Data matrix $\mathbf{X}$, the adjacency matrix of an incomplete graph structure $\mathbf{A}$, and parameters $\lambda$ and $d_v$.
\end{algorithmic}
{\bfseries Initialize:}
$epochs = 800$;
\begin{algorithmic}[1]
\FOR{ ${t = 1}$ to $epochs$ }
\STATE Constructing two complementary augmented views $\mathbf{A_1}$ and $\mathbf{A_2}$ from $\mathbf{A}$;
\FOR{ ${v = 1}$ to $2$ }
\STATE {$i = 1$} {\bfseries if} {$v == 2$} {\bfseries else} 2;
\STATE Computing ${\mathbf{Z}^{(v)}}$ via Eq. \eqref{eq:fz} using $\mathbf{X}$ and $\mathbf{A}_v$;
\STATE Constructing the set $\mathbf{E}_i^{'}$ by sampling the negative edges from the other augmented view $\mathbf{A}_i$;
\STATE Computing ${\mathbf{H}^{(v)}}$ via Eq. \eqref{eq:fip} using three variables, including ${\mathbf{Z}^{(v)}}$, the set of edges constructed from $\mathbf{A}_i$, and the set of negative edges $\mathbf{E}_i^{'}$;
\STATE Computing $\widetilde {\mathbf{A}_i}$ via Eq. \eqref{eq:fmlp}; \\
\STATE Updating ${{\mathbf{W}^{(1)}}}$ and ${{\mathbf{W}^{(1,2)}}}$ by minimizing ${L_{v}}$ in Eq. \eqref{eq:lossr} using $\mathbf{A}_i$ and $\widetilde {\mathbf{A}_i}$; \\
\ENDFOR
\ENDFOR
\STATE Computing ${\mathbf{Z}}$ via Eq. \eqref{eq:fz} using $\mathbf{X}$ and $\mathbf{A}$;\\
\STATE Computing ${\mathbf{H}}$ via Eq. \eqref{eq:fip}; \\
\STATE  Computing $\widetilde {\mathbf{A}}$ via Eq. \eqref{eq:fmlp}; \\
\ENSURE The graph structure $\widetilde {\mathbf{A}}$.
\end{algorithmic}
\end{algorithm}

Similarly, the sufficient supervisory information shared between one augmented view and the graph structure reconstructed from the other augmented view is task-relevant if the following condition holds, i.e.,
\begin{equation}\label{eq:reltask}
\begin{split}
C\left( {{\mathbf{A}_1},\widetilde {\mathbf{A}_1}} \right) = C\left( {{\mathbf{A}_2},\widetilde {\mathbf{A}_2}} \right) = C\left( {\widetilde {\mathbf{A}_1},\widetilde {\mathbf{A}_2}} \right).
\end{split}
\end{equation}

By considering the objective from the perspective of sufficient supervisory information, we emphasize the importance of utilizing appropriate data augmentation schemes for incomplete graph structures. Therefore, the augmented view generation process can typically be guided by the theoretical conditions in \eqref{eq:infoloss} and \eqref{eq:reltask}.

\begin{theorem} \label{th:bound}
For any two variables $r_1 \in \left[ {0,1} \right]$ and $r_2 \in \left[ {0,1} \right]$, $C\left( {{a_1},{a_2}} \right)$ is bounded by:
\begin{equation*}
 - \frac{{\log 2}}{{1 + {1 \mathord{\left/
 {\vphantom {1 e}} \right.
 \kern-\nulldelimiterspace} e}}} \le C\left( {{a_1},{a_2}} \right) \le  - \frac{{\log \left( {1 + {1 \mathord{\left/
 {\vphantom {1 e}} \right.
 \kern-\nulldelimiterspace} e}} \right)}}{2}
\end{equation*}
\end{theorem}
where ${a_1} = \delta \left( {{r_1}} \right)$, ${a_2} = \delta \left( {{r_2}} \right)$, and $\delta \left(  \cdot  \right)$ denotes the sigmoid function.

\begin{proof}
Let $r_i \in \left[ {0,1} \right]$ $\left( i \in \left\{ {1,2} \right\} \right)$ be any two variables. According to the sigmoid function $\delta \left(  \cdot  \right)$, we have
\begin{equation*}
\frac{1}{2} \le {a_i} \le \frac{1}{{1 + {1 \mathord{\left/
 {\vphantom {1 e}} \right.
 \kern-\nulldelimiterspace} e}}}.
\end{equation*}
Considering $C\left( {{a_1},{a_2}} \right) = {\mathbb{E}_{p_{\left({a_1},{a_2}\right)}}}{a_1}\log {a_2}$, we obtain
\begin{equation*}
 - \frac{{\log 2}}{{1 + {1 \mathord{\left/
 {\vphantom {1 e}} \right.
 \kern-\nulldelimiterspace} e}}} \le C\left( {{a_1},{a_2}} \right) \le  - \frac{{\log \left( {1 + {1 \mathord{\left/
 {\vphantom {1 e}} \right.
 \kern-\nulldelimiterspace} e}} \right)}}{2}.
\end{equation*}
\QEDB
\end{proof}

According to Theorem \ref{th:bound}, ${L_v}$ has specific upper and lower bounds in Eq. \eqref{eq:lossr}. The lower bound can be theoretically guaranteed when minimizing ${L_v}$ in Eq. \eqref{eq:lossr}. Moreover, we further obtain the following Lemma.
\begin{lemma} \label{lm:bound}
There exists a constant such that $\forall a_1, a_2$ and $\forall {\theta _1}, {\theta _2} \in \theta $, the following inequality holds:
\begin{equation*}
\left| {{C_{{\theta _1}}}\left( {{a_1},{a_2}} \right) - {C_{{\theta _2}}}\left( {{a_1},{a_2}} \right)} \right| \le \frac{1}{{1 + {1 \mathord{\left/
 {\vphantom {1 e}} \right.
 \kern-\nulldelimiterspace} e}}}
 \end{equation*}
where $\theta \in {\mathbb{R}^d}$ represents the parameters in neural networks.
\end{lemma}
The proof of Lemma \ref{lm:bound} is omitted, as it follows in a straightforward manner from Theorem \ref{th:bound}. Lemma \ref{lm:bound} predicts that the loss function ${L}_v$ will gradually decline during the training stage.

\section{Related Work}
\label{sec:wrok}
Contrastive learning-based GRL methods follow the principle of mutual information maximization by contrasting positive and negative pairs \cite{You2020GCL, Hassani2020CMRL, Zhu2020GCRL}. Data augmentation is a key prerequisite for these GRL methods. For example, You \textit{et al.} \cite{You2020GCL} presented a graph-contrastive learning framework that provides four types of graph augmentation strategies, including node dropping, edge perturbation, attribute masking and subgraph production, to improve the generalizability of the graph representations produced during GNN pretraining.

Generative learning-based GRL methods aim to reconstruct graph data for learning graph representations \cite{Tan2023GAE, Li2022MGAE, Kipf2016VGAE}. For example, a VGAE employs two GCN models to build its encoder component, which can learn meaningful node embeddings for the reconstruction of a graph structure \cite{Kipf2016VGAE}. Lousi \textit{et al.} \cite{Louis2023S3GRL} presented a scalable simplified subgraph representation learning method (S3GRL) that simplifies the message passing and aggregation operations in the subgraph of each link. Additionally, several GAE-based GRL methods employ different masking strategies on a graph structure to implement graph augmentation, including a masked GAE (GraphMAE) \cite{Hou2022SMGA}, an MGAE \cite{Li2022MGAE} and a self-supervised GAE (S2GAE) \cite{Tan2023GAE}. After randomly making a portion of the edges on a graph structure, these GRL methods attempt to reconstruct the missing connections with a partially visible, unmasked graph structure.

\section{Experiments}
\label{sec:exp}
In this section, we conduct extensive experiments to evaluate the link prediction performance of the proposed CGCL method. The source code for CGCL is implemented upon a PyTorch framework and a PyTorch Geometric (PyG) library. All experiments are performed on a Linux workstation with a GeForce RTX 4090 GPU (24-GB caches), an Intel (R) Xeon (R) Platinum 8336C CPU and 128.0 GB of RAM.

\begin{table*}[!htbp]
\scriptsize
\setlength{\tabcolsep}{2pt}
\centering
\begin{tabular}{c|c|cc|cc|cc|cc|cc}
\hline
\multirow{2}*{Methods} & \multirow{2}*{Testing ratios} & \multicolumn{2}{c|}{Cora} &  \multicolumn{2}{c|}{CiteSeer} & \multicolumn{2}{c|}{PubMed} & \multicolumn{2}{c|}{Photo} & \multicolumn{2}{c}{Computers} \\
\cline{3-12}
~ & ~ & AUC & AP & AUC & AP & AUC & AP & AUC & AP & AUC & AP \\
\hline
GraphSAGE &  \multirow{7}*{10\%}  & 90.56$\pm$0.39  & 88.70$\pm$0.62 & 89.22$\pm$0.47 & 88.15$\pm$0.67 & 94.85$\pm$0.10 & 94.97$\pm$0.08 & 90.42$\pm$1.35 & 87.99$\pm$1.27 & 87.20$\pm$1.38 & 84.99$\pm$1.18 \\
VGAE & ~ & 91.88$\pm$0.58 & 92.51$\pm$0.35 & 91.29$\pm$0.56 & 92.35$\pm$0.41 & 94.86$\pm$0.69 & 94.75$\pm$0.62 & 96.76$\pm$0.16 & 96.72$\pm$0.16 & 90.86$\pm$0.69 & 91.75$\pm$0.62 \\
SEAL & ~ & 91.11$\pm$0.13  & 92.87$\pm$0.15 & 89.08$\pm$0.15 & 91.44$\pm$0.05 & 93.18$\pm$0.17 & 93.84$\pm$0.07 & 96.46$\pm$0.14  & 97.11$\pm$0.13 & 94.53$\pm$0.16 & 94.69$\pm$0.31 \\
S3GRL & ~ & 94.24$\pm$0.22 & 94.04$\pm$0.54 & 95.79$\pm$0.63 & 95.04$\pm$0.68 & 97.12$\pm$0.29 & 96.72$\pm$0.40 & 96.90$\pm$0.52 & 96.44$\pm$0.43 & 95.17$\pm$0.45 & 93.08$\pm$0.65 \\
S2GAE & ~ & 95.89$\pm$0.48  & 95.78$\pm$0.60 & 95.65$\pm$0.26 & 95.75$\pm$0.33 & 96.85$\pm$0.13 & 96.49$\pm$0.13 & 97.93$\pm$0.06 & 97.39$\pm$0.09 & 97.25$\pm$0.17 & 96.68$\pm$0.21 \\
MGAE & ~ & \underline{96.74 $\pm$0.09}  & \underline{96.36 $\pm$0.18} & \textbf{97.62$\pm$0.13} & \textbf{97.87$\pm$0.13} & \underline{97.62$\pm$0.02} & \underline{97.19$\pm$0.04} & \underline{98.64$\pm$0.01} & \underline{98.42$\pm$0.02} & \underline{98.27$\pm$0.03} & \underline{98.01$\pm$0.03} \\
CGCL & ~ & \textbf{97.00$\pm$0.15} & \textbf{97.34$\pm$0.11} & \underline{97.35$\pm$0.23} & \underline{97.62$\pm$0.16} & \textbf{98.48$\pm$0.03} & \textbf{98.37$\pm$0.03} & \textbf{98.88$\pm$0.01} & \textbf{98.72$\pm$0.02} & \textbf{98.41$\pm$0.04} & \textbf{98.15$\pm$0.04} \\
\hline
GraphSAGE &  \multirow{7}*{20\%}  & 90.17$\pm$0.60  & 88.62$\pm$0.67 & 88.09$\pm$0.74 & 85.62$\pm$0.84 & 94.26$\pm$0.14 & 94.44$\pm$0.16 & 89.58$\pm$2.68 & 87.45$\pm$3.28 & 86.04$\pm$1.19 & 84.25$\pm$1.55 \\
VGAE & ~ & 89.73$\pm$0.24  & 91.03$\pm$0.26 & 90.66$\pm$0.38 & 91.59$\pm$0.36 & 93.44$\pm$0.4 & 93.71$\pm$0.26 & 96.61$\pm$0.16 & 96.42$\pm$0.16 & 89.44$\pm$0.40 & 90.71$\pm$0.26 \\
SEAL & ~ & 91.40$\pm$0.22  & 92.80$\pm$0.13 & 89.13$\pm$0.23 & 91.33$\pm$0.06 & 93.22$\pm$0.16 & 93.90$\pm$0.13 & 94.68$\pm$0.09 & 95.26$\pm$0.14 & 90.13$\pm$0.23 & 91.46$\pm$0.27 \\
S3GRL & ~ & 92.18$\pm$0.32 & 92.01$\pm$0.58 & 94.76$\pm$0.56 & 94.42$\pm$0.68 & 95.55$\pm$0.52 & 95.16$\pm$0.71 & 95.32$\pm$0.27 & 95.17$\pm$0.94 & 92.85$\pm$0.58 & 93.94$\pm$0.77 \\
S2GAE & ~ & 93.00$\pm$0.37  & 92.42$\pm$0.60 & 94.82$\pm$0.22 & 94.74$\pm$0.26 &  96.60$\pm$0.15 & 96.24$\pm$0.15 & 97.76$\pm$0.09 & 97.19$\pm$0.13  & 97.21$\pm$0.10 & 96.65$\pm$0.13 \\
MGAE & ~ & \underline{95.75$\pm$0.11} & \underline{96.23$\pm$0.09} & \underline{95.75$\pm$0.11} & \underline{96.23$\pm$0.09} & \underline{97.47$\pm$0.03} & \underline{97.20$\pm$0.03} & \underline{98.51$\pm$0.01} & \underline{98.29$\pm$0.02} & \underline{98.11$\pm$0.03} & \underline{97.88$\pm$0.03} \\
CGCL & ~ & \textbf{96.61$\pm$0.35} & \textbf{97.16$\pm$0.15} & \textbf{97.21$\pm$0.18} & \textbf{97.51$\pm$0.13} & \textbf{98.27$\pm$0.02} & \textbf{98.14$\pm$0.04} & \textbf{98.77$\pm$0.01} & \textbf{98.59$\pm$0.01} & \textbf{98.32$\pm$0.02} & \textbf{98.11$\pm$0.03} \\
\hline
\end{tabular}
\caption{Link prediction results obtained on all datasets.}
\label{tb:prediction:results}
\end{table*}

\begin{table*}[!htbp]
\scriptsize
\setlength{\tabcolsep}{2pt}
\centering
\begin{tabular}{c|c|cc|cc|cc|cc|cc}
\hline
\multirow{2}*{Methods} & \multirow{2}*{Testing ratios} & \multicolumn{2}{c|}{Cora} &  \multicolumn{2}{c|}{CiteSeer} & \multicolumn{2}{c|}{PubMed} & \multicolumn{2}{c|}{Photo} & \multicolumn{2}{c}{Computers} \\
\cline{3-12}
~ & ~ & AUC & AP & AUC & AP & AUC & AP & AUC & AP & AUC & AP \\
\hline
DGCL$_{\textbf{one-view}}$ & \multirow{2}*{10\%} & 96.77$\pm$0.13 & 97.13$\pm$0.12 & 97.10$\pm$0.17  & 97.20$\pm$0.19 & 98.30$\pm$0.03 & 98.21$\pm$0.04 & 98.63$\pm$0.04 & 98.43$\pm$0.05 & 98.08$\pm$0.03 & 97.96$\pm$0.04 \\
CGCL & ~ & \textbf{97.00$\pm$0.15} & \textbf{97.34$\pm$0.11} & \textbf{97.26$\pm$0.23} & \textbf{97.54$\pm$0.16} & \textbf{98.48$\pm$0.03} & \textbf{98.37$\pm$0.03} & \textbf{98.88$\pm$0.01} & \textbf{98.72$\pm$0.02} & \textbf{98.39$\pm$0.04} & \textbf{98.41$\pm$0.04} \\
\hline
DGCL$_{\textbf{one-view}}$ & \multirow{2}*{20\%} & 96.91$\pm$0.13 & 96.95$\pm$0.12 & 97.09$\pm$0.12 & 97.18$\pm$0.14 & 98.19$\pm$0.03 & 98.02$\pm$0.04 & 98.61$\pm$0.05 & 98.45$\pm$0.04 & 98.01$\pm$0.03 & 97.91$\pm$0.05 \\
CGCL & ~ & \textbf{97.61$\pm$0.35} & \textbf{97.16$\pm$0.15} & \textbf{97.28$\pm$0.18} & \textbf{97.51$\pm$0.13}& \textbf{98.27$\pm$0.02} & \textbf{98.14$\pm$0.04} & \textbf{98.77$\pm$0.01} & \textbf{98.59$\pm$0.01} & \textbf{98.32$\pm$0.02} & \textbf{98.37$\pm$0.03} \\
\hline
\end{tabular}
\caption{Ablation study concerning the main training stages of the proposed CGCL method conducted on all datasets.}
\label{tb:ablation:results}
\end{table*}

\subsection{Experimental Settings}
\subsubsection{Datasets}
We select five widely used graph datasets for evaluation, including Cora \cite{Sen2008CC}, Citeseer \cite{Sen2008CC}, Pubmed \cite{Namata2012QCS}, Photo \cite{McAuley2015IR}, and Computers \cite{McAuley2015IR}, which are publicly available on PyG. The Cora, Citeseer and Pubmed datasets are citation networks, where nodes and edges indicate papers and citations, respectively. The Photo and Computers datasets are segments of the Amazon co-purchase graph, where each node represents a good, and each edge indicates that the two corresponding goods are frequently bought together.

Each graph dataset is divided into three parts, including a training set, a validation set and a testing set. We use two different sets of percentages for the validation set and testing set, including (1) 5\% of the validation set and 10\% of the testing set and (2) 10\% of the validation set and 20\% of the testing set. The links in the validation and testing sets are masked in the training graph structure. For example, we randomly select 5\% and 10\% of the links and the same numbers of disconnected node pairs as testing and validation sets under the first setting, respectively. The remainder of the links in the graph structure are used for training.

\subsubsection{Comparison Methods}
We compare the proposed CGCL method with several state-of-the-art methods for link prediction, including GraphSAGE \cite{Hamilton2017SAGE}, a VGAE \cite{Veli2018GAT}, SEAL \cite{Zhang2021SEAL}, S3GRL \cite{Louis2023S3GRL}, S2GAE \cite{Tan2023GAE}, and an MGAE \cite{Li2022MGAE}. The source codes of the competing algorithms are provided by their respective authors. For the MGAE, the edgewise random masking strategy is chosen to sample a subset of the edges in each dataset.

\subsubsection{Evaluation Metrics}
Two metrics are utilized to evaluate the link prediction performance of all competing algorithms, including the area under the curve (AUC) score and average precision (AP) score. For comparison, each experiment is conducted 10 times with different random parameter initializations. We report the mean values and standard deviations achieved by all the competing methods on the five graph datasets. For each evaluation metric, a higher value represents better link prediction performance.

\subsubsection{Parameter Settings}
\label{sec:settings}
The proposed network architecture contains 2 hidden layers in the CGCL model. The sizes of the 2 hidden layers are set to $\left[ {{d_v}, {{{d_v}} \mathord{\left/
 {\vphantom {{{d_v}} 4}} \right.
 \kern-\nulldelimiterspace} 2}} \right]$, where ${d_v}$ is the number of neural units in the first hidden layer. In the experiments, ${d_v}$ ranges within $\left\{ 512, 256, 128, 64 \right\}$. The learning rate of the proposed CGCL method $r$ is chosen from $\left\{1{e^{ - 3}}, 5{e^{ - 3}}, 0.01, 0.05 \right\}$. For all datasets, the number of iterations is set to 800 during the training stage. To conduct a fair comparison, the best link prediction results of these competing methods are obtained by tuning their parameters.

\subsection{Performance Evaluation}
\label{sec:evaluation}
The experimental results produced by all competing methods on the five link prediction tasks are reported in Table \ref{tb:prediction:results}. The best and second-best values of the link prediction results are highlighted in bold and underlined, respectively. We observed that the proposed CGCL method almost performs better than the other competing methods in terms of the AUC and AP. For example, CGCL achieves performance improvements of approximately 0.26\%, 0.86\%, 0.24\% and 0.13\% in terms of the AUC with a testing rate of 10\% on the Cora, PubMed, Photo and Computers datasets, respectively. Moreover, CGCL outperforms the competing methods in terms of the AUC and AP metrics as the testing rate increases from 10\% to 20\% in the link prediction tasks. These results demonstrate the superiority of CGCL over the other methods.

\begin{figure}
    \centering
    \subfloat[Testing ratio = 10\%]{\includegraphics[width=0.48\linewidth]{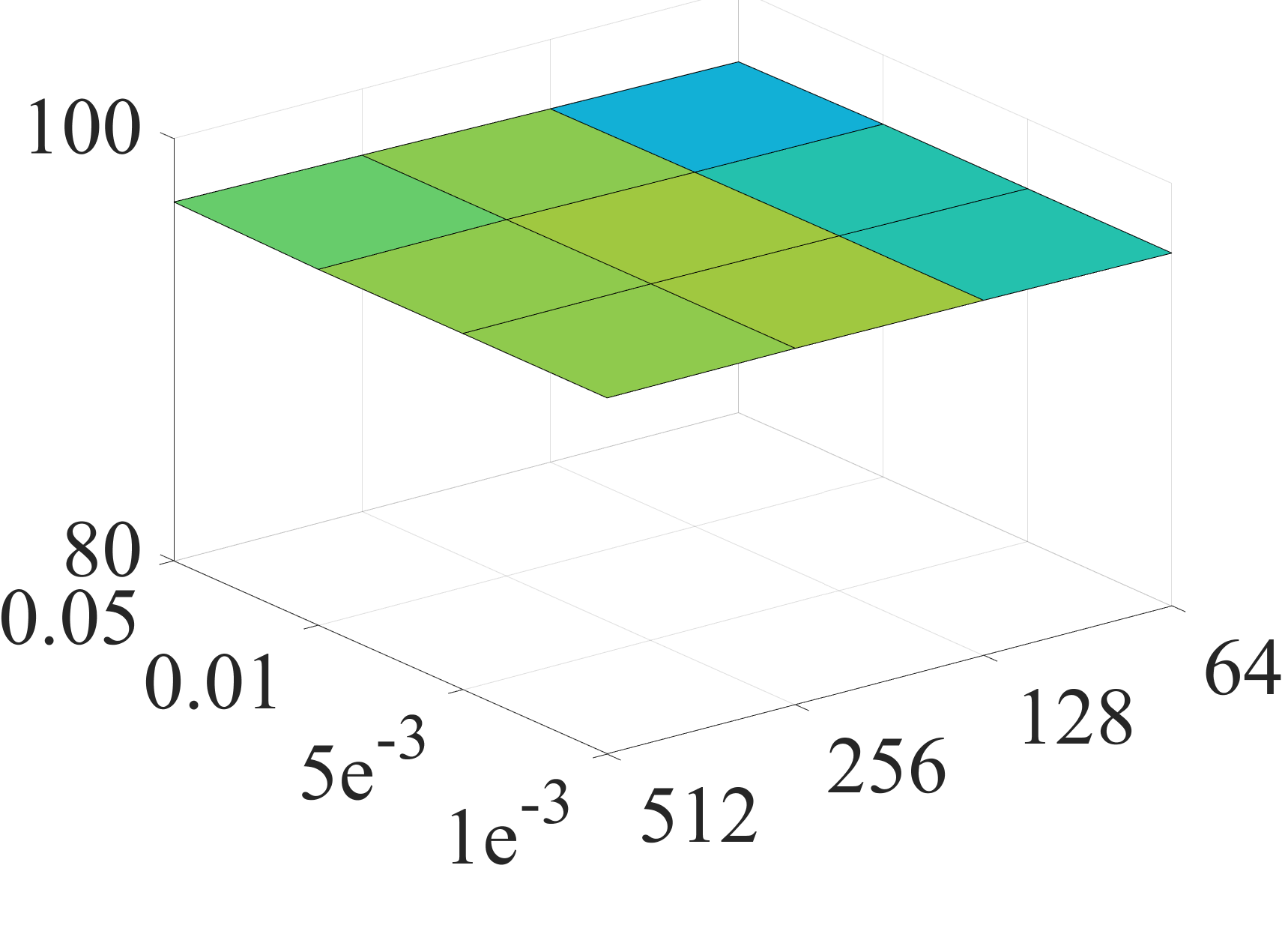}%
    \label{fig:parameters:cora:a}}
    \hfil
    \subfloat[Testing ratio = 10\%]{\includegraphics[width=0.48\linewidth]{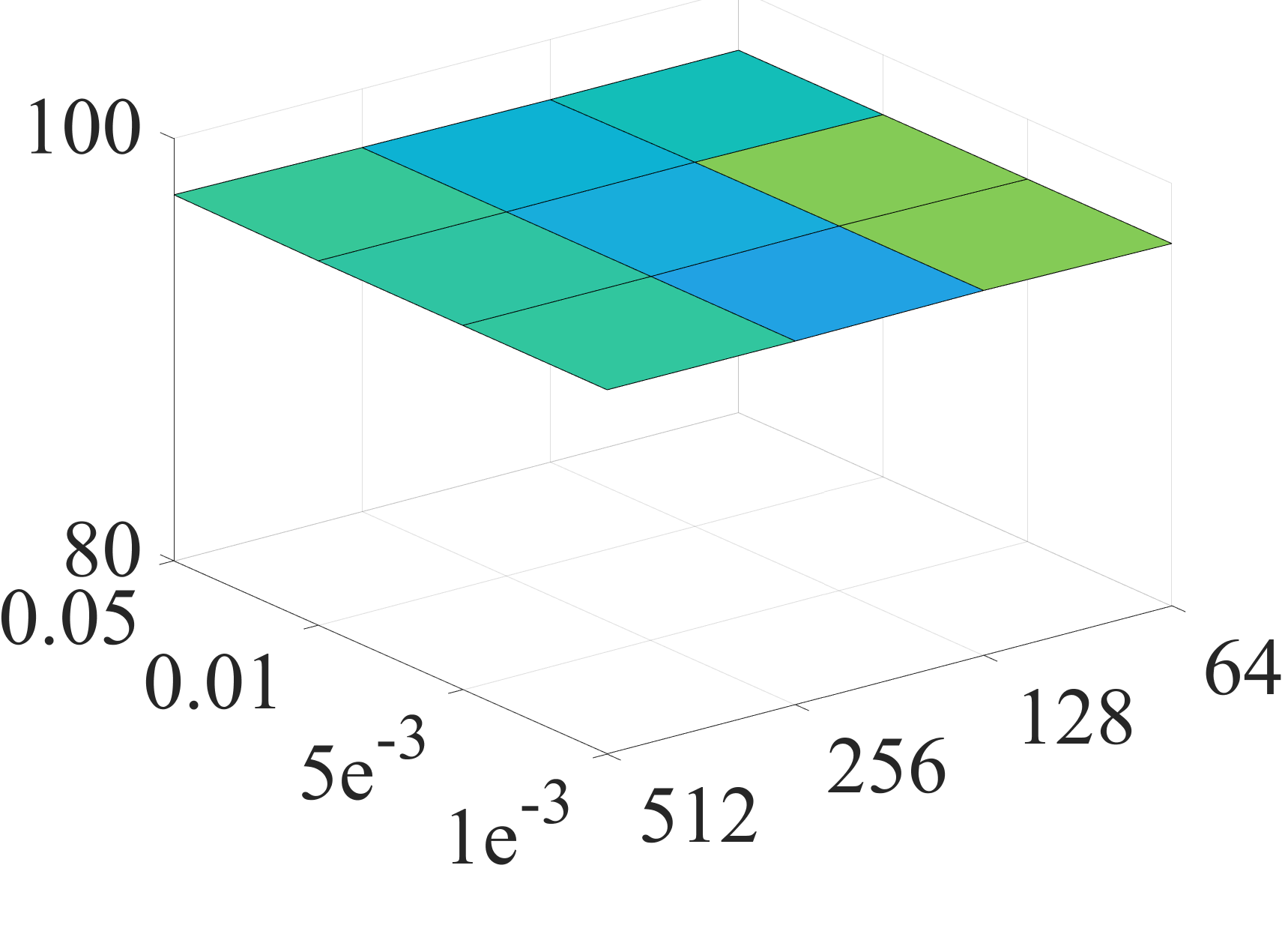}%
    \label{fig:parameters:cora:b}}

    \subfloat[Testing ratio = 20\%]{\includegraphics[width=0.48\linewidth]{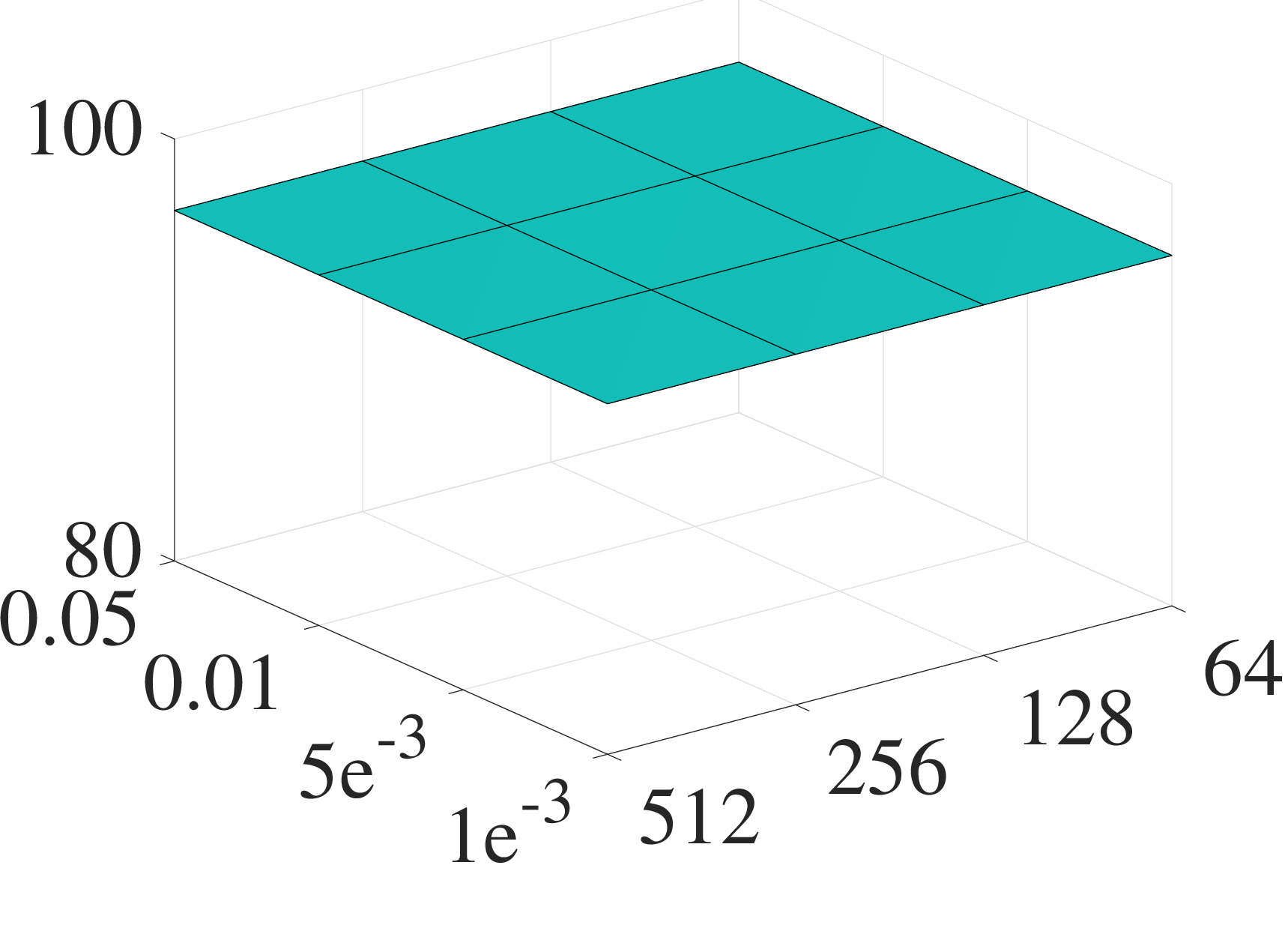}%
    \label{fig:parameters:cora:c}}
    \hfil
    \subfloat[Testing ratio = 20\%]{\includegraphics[width=0.48\linewidth]{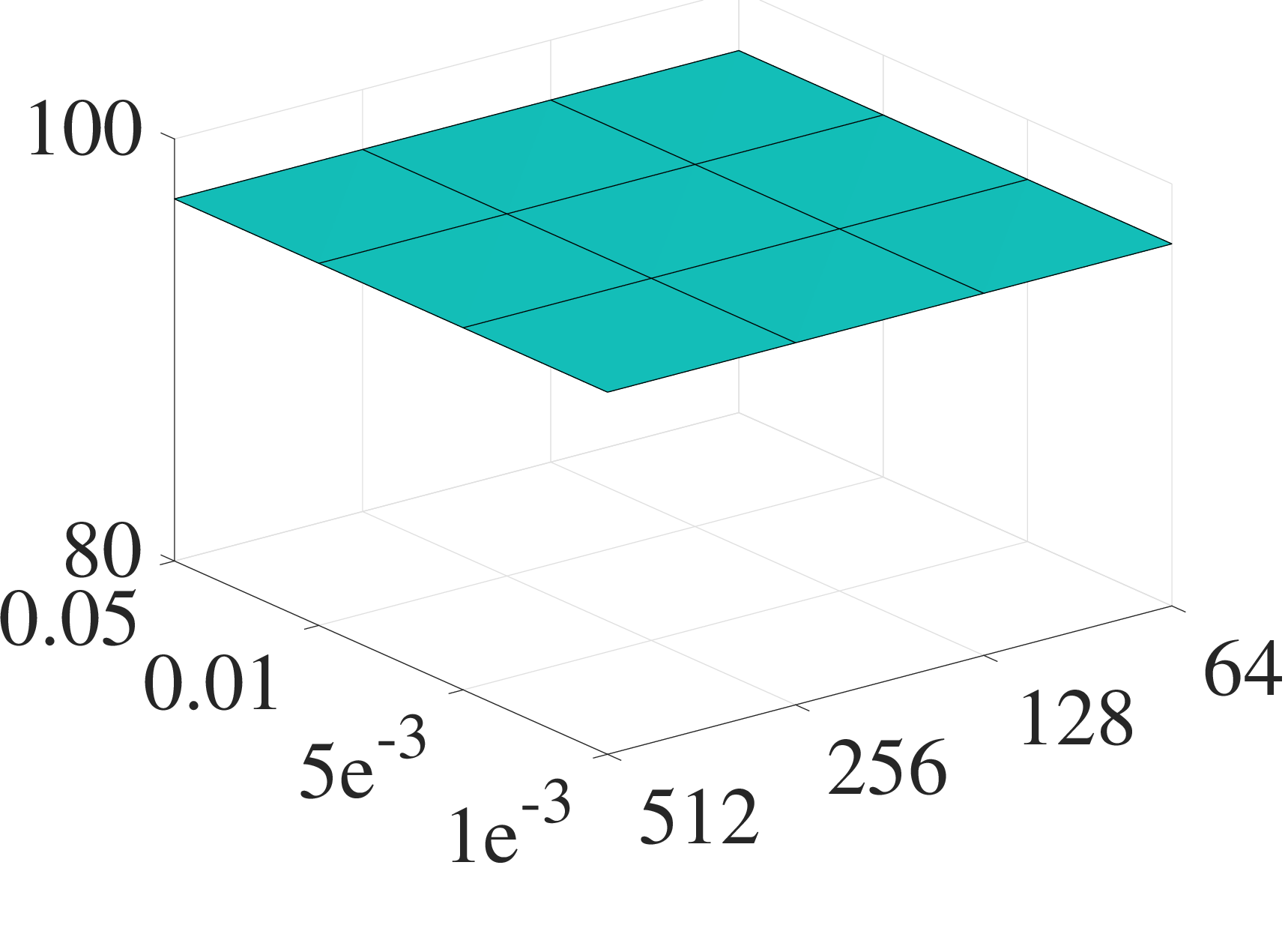}%
    \label{fig:parameters:cora:d}}
    \caption{The AUC and AP values yielded by the CGCL method with different combinations of $d_v$ and $r$ on the Cora dataset.}
    \label{fig:parameters:cora}
\end{figure}

As expected, the AUC and AP results produced by CGCL slightly decrease as the testing rate increases from 10\% to 20\% on the three citation datasets, including the Cora, Citeseer and PubMed datasets. In contrast, the AUC and AP results of CGCL remain almost unchanged on the Photo and Computers datasets. These two datasets contain larger numbers of edges than the other datasets. This demonstrates the effectiveness and robustness of the coupled graph structure augmentation approach.

Two main reasons highlight the advantages of the proposed CGCL method. First, constructing two complementary augmented views of a graph structure enhances the diversity of the produced graph representations while counteracting the adverse consequences of information losses in the graph representations. Additionally, the MGAE yields encouraging AUC and AP results in the experiments due to its effective edge masking strategy. Second, the proposed CGCL method achieves invariant graph representations through cross-view graph consistency learning. This mechanism plays a key role in facilitating the reconstruction of the incomplete graph structure within the scope of self-supervised learning.

\begin{figure}
    \centering
    \subfloat[Testing ratio = 10\%]{\includegraphics[width=0.48\linewidth]{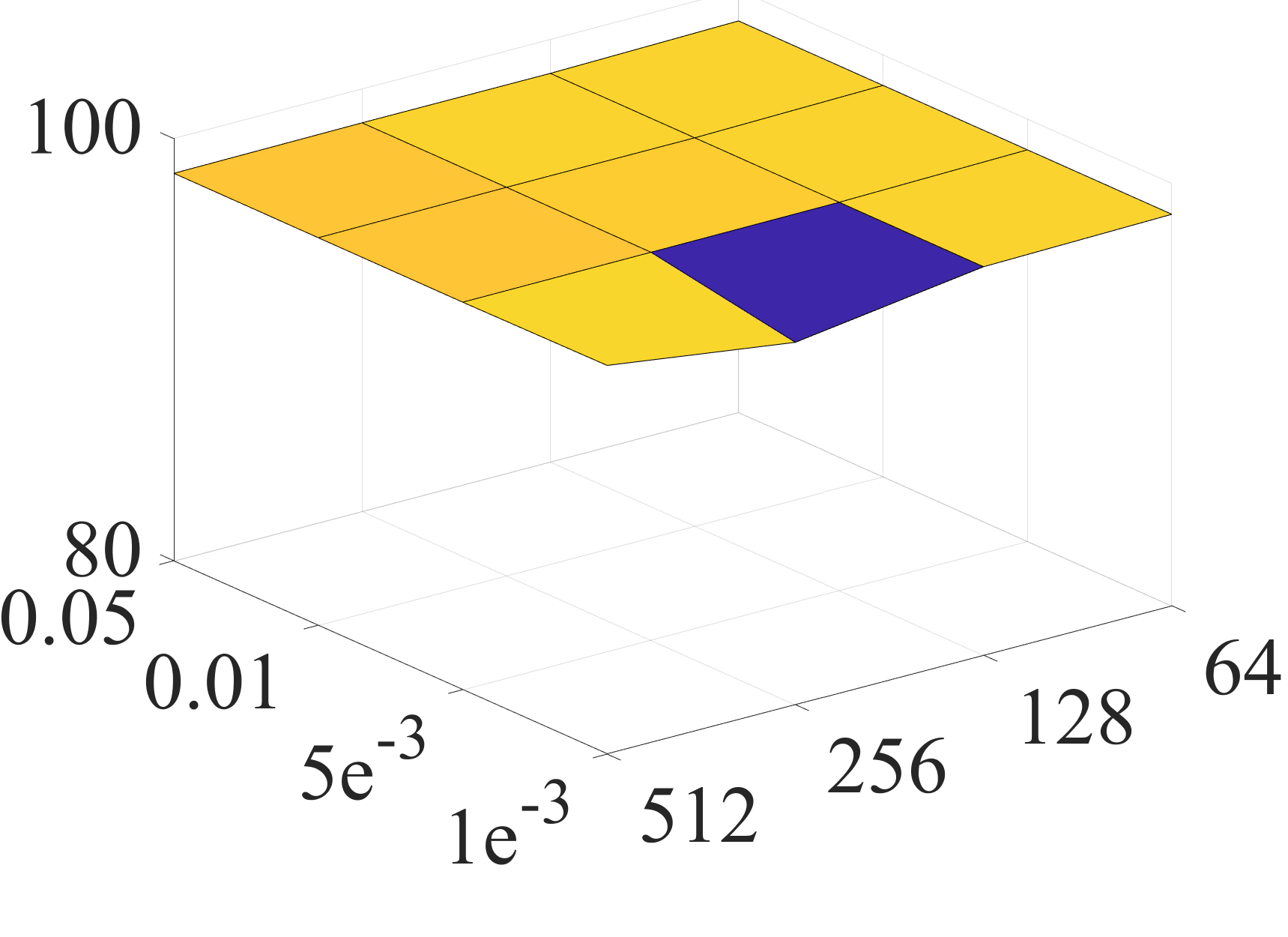}%
    \label{fig:parameters:computers:a}}
    \hfil
    \subfloat[Testing ratio = 10\%]{\includegraphics[width=0.48\linewidth]{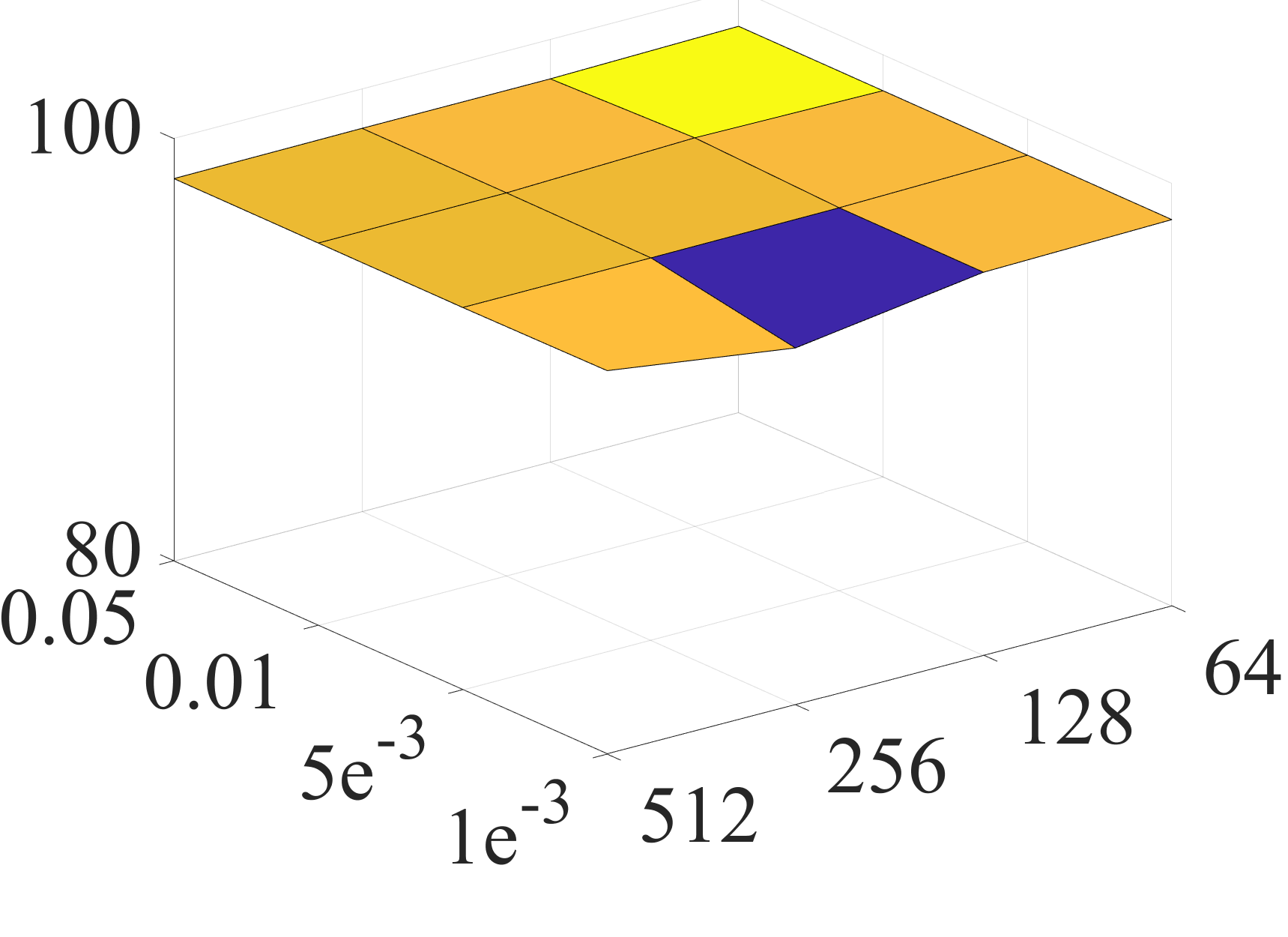}%
    \label{fig:parameters:computers:b}}

    \subfloat[Testing ratio = 20\%]{\includegraphics[width=0.48\linewidth]{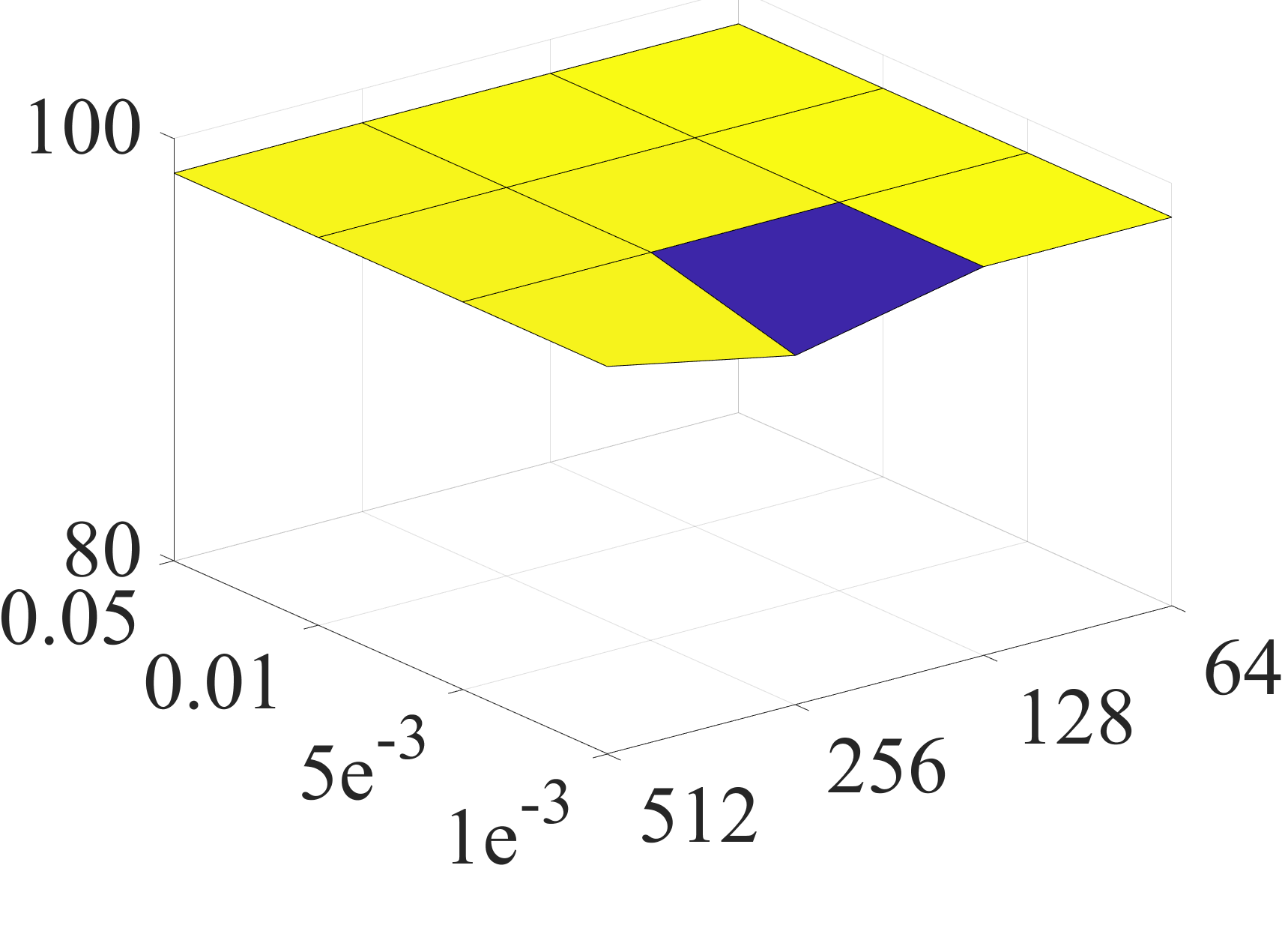}%
    \label{fig:parameters:computers:c}}
    \hfil
    \subfloat[Testing ratio = 20\%]{\includegraphics[width=0.48\linewidth]{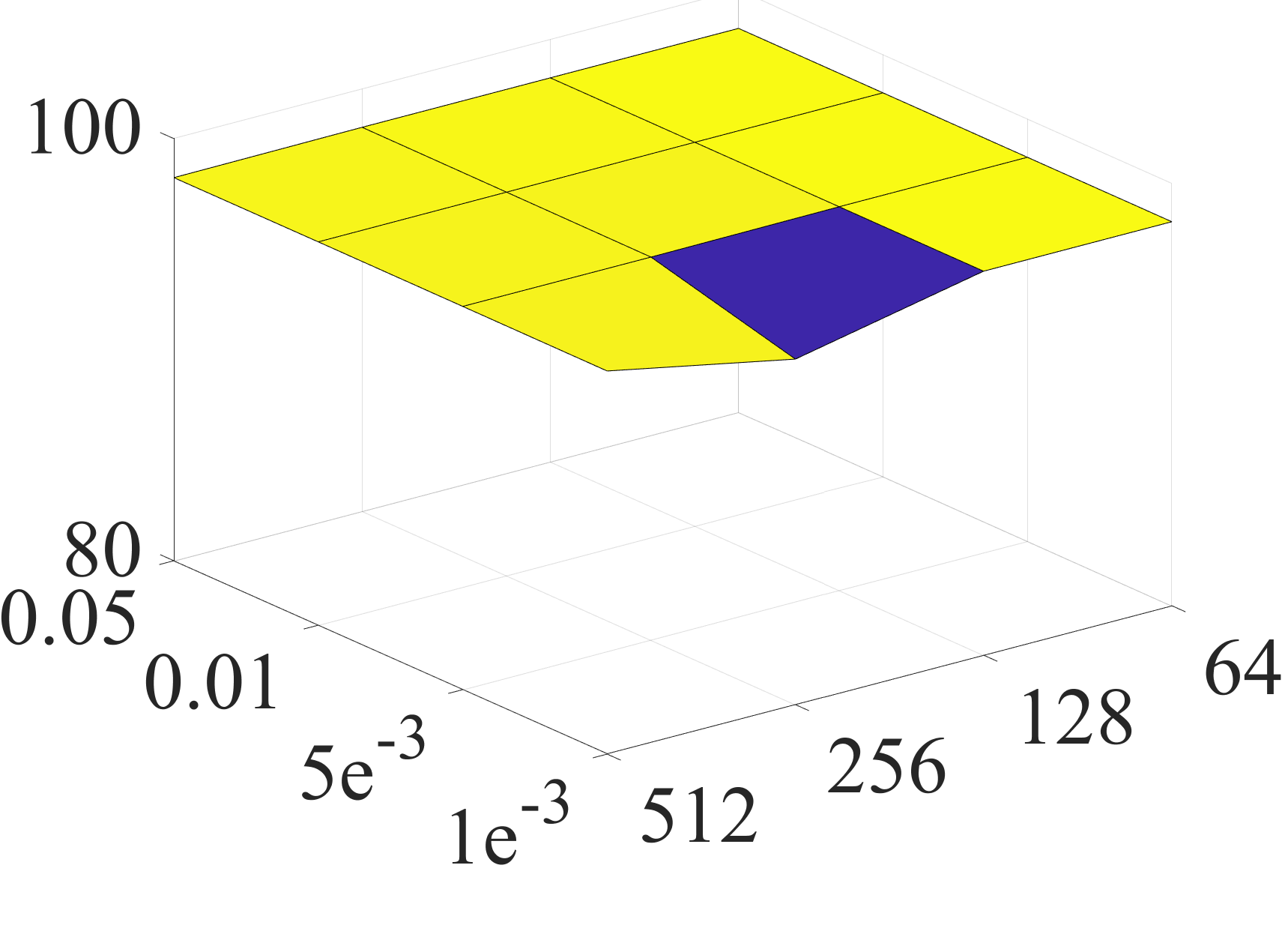}%
    \label{fig:parameters:computers:d}}
    \caption{The AUC and AP values yielded by the CGCL method with different combinations of $d_v$ and $r$ on the Computers dataset.}
    \label{fig:parameters:computers}
\end{figure}

\begin{figure}
    \centering
    \subfloat[Testing ratio = 10\%]{\includegraphics[width=0.48\linewidth]{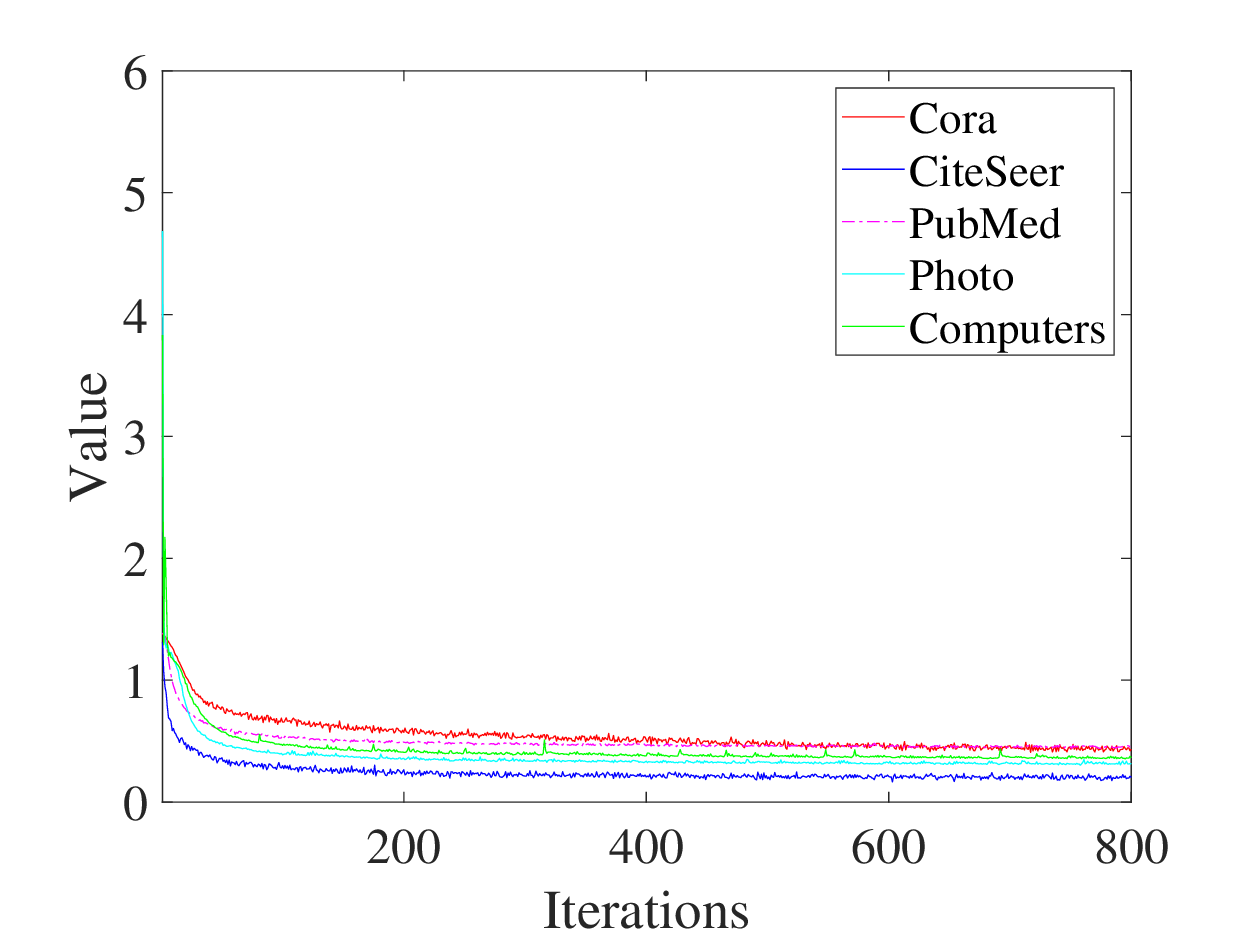}%
    \label{fig:convergence:a}}
    \hfil
    \subfloat[Testing ratio = 20\%]{\includegraphics[width=0.48\linewidth]{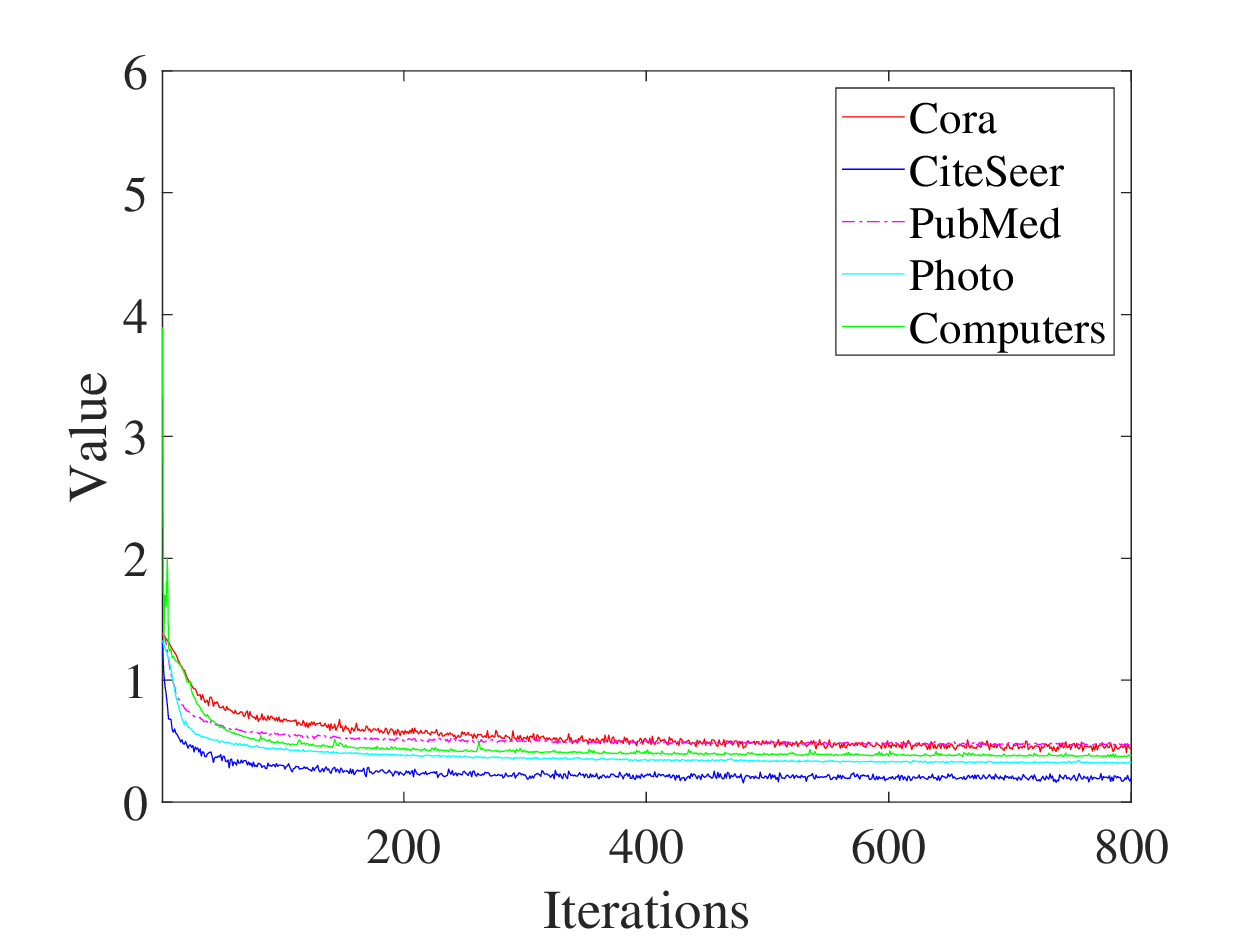}%
    \label{fig:convergence:b}}
    \caption{Convergence results obtained by the CGCL method on all the datasets.}
    \label{fig:convergence}
\end{figure}

\subsection{Ablation Study}
\label{sec:ablation}
In Section \ref{sec:training}, CGCL constructs two complementary augmented graph structure views using the coupled graph structure augmentation scheme. To verify the importance of the proposed graph structure augmentation scheme in CGCL, we further conduct an ablation study to isolate the necessity of the two complementary augmented views. Specifically, we consider a special version, i.e., a variant that only chooses one of the two complementary augmented views during the training stage, which is referred to as DGCL$_{\textbf{one-view}}$. We employ the same experimental settings as those utilized above. The best experimental results derived from the two complementary augmented views are included for comparison purposes. Table \ref{tb:ablation:results} shows the experimental AUC and AP results produced by DGCL$_{\textbf{one-view}}$ and CGCL. We see that CGCL performs better than DGCL$_{\textbf{one-view}}$ on the link prediction tasks. This provides strong empirical evidence demonstrating the importance of the two complementary augmented views in CGCL.

\subsection{Parameter Sensitivity Study}
We conduct experiments to investigate the sensitivity levels of the two parameters in the proposed CGCL method, including the number of neural units in the first hidden layer $d_v$ and the learning rate $r$. The $d_v$ and $r$ parameters range within $\left\{ 512, 256, 128, 64 \right\}$ and $\left\{1{e^{ - 3}}, 5{e^{ - 3}}, 0.01, 0.05 \right\}$, respectively. Due to space limitations, two representative datasets are selected for evaluation, i.e., the Cora and Computers datasets. Figs. \ref{fig:parameters:cora} and \ref{fig:parameters:computers} show the experimental results achieved by CGCL in terms of the AUC and AP values obtained with different combinations of $d_v$ and $r$. CGCL can achieve relatively stable link prediction results under different testing ratios with different combinations of $d_v$ and $r$. This indicates that CGCL performs well across relatively large $d_v$ and $r$ ranges on the Cora and Computers datasets.

\section{Conclusion}
\label{sec:conclusion}
In this paper, we present a CGCL method that learns invariant graph representations from graph-structured data. CGCL utilizes a coupled graph structure augmentation scheme to construct two complementary augmented graph structure views. This augmentation scheme supports graph consistency learning, thereby enhancing the generalizability of the graph representations produced by CGCL. Through cross-view graph consistency learning, CGCL effectively acquires invariant graph representations and facilitates the construction of an incomplete graph structure. Extensive experimental results obtained on graph datasets demonstrate that the proposed CGCL method almost outperforms several state-of-the-art approaches.

\section{Acknowledgments}
This work was supported in part by National Natural Science Foundation of China (NSFC) under Grants U21B2040, 62176171 and 72374089, in part by the Fundamental Research Funds for the Central Universities under Grant CJ202303, and in part by Sichuan Science and Technology Planning Project under Grant 24NSFTD0130.

\end{document}